\documentclass{article}

\usepackage{PRIMEarxiv}

\usepackage[utf8]{inputenc} 
\usepackage[T1]{fontenc}    
\usepackage{hyperref}       
\usepackage{url}            
\usepackage{booktabs}       
\usepackage{amsfonts}       
\usepackage{nicefrac}       
\usepackage{microtype}      
\usepackage{lipsum}
\usepackage{fancyhdr}       
\usepackage{graphicx}       
\graphicspath{{media/}}     

\usepackage{makecell}
\usepackage{multirow}
\usepackage{natbib}
\setcitestyle{authoryear,open={(},close={)}}

\title{Open-source Frame Semantic Parsing}

\author{
  David Chanin \\
  Department of Computer Science \\
  University College London \\
  \texttt{david.chanin.22@ucl.ac.uk} \\
}

\begin{document}
\maketitle

\begin{abstract}
While the state-of-the-art for frame semantic parsing has progressed dramatically in recent years, it is still difficult for end-users to apply state-of-the-art models in practice. To address this, we present Frame Semantic Transformer, an open-source Python library which achieves near state-of-the-art performance on FrameNet 1.7, while focusing on ease-of-use. We use a T5 model fine-tuned on Propbank and FrameNet exemplars as a base, and improve performance by using FrameNet lexical units to provide hints to T5 at inference time. We enhance robustness to real-world data by using textual data augmentations during training.
\end{abstract}

\section{Introduction}

Frame semantic parsing \citep{gildea2002automatic} is a natural language understanding (NLU) task involving finding structured semantic frames and their arguments from natural language text as formalized by the FrameNet project \citep{baker1998berkeley}. Frame semantics has proved useful in understanding user intent from text, finding use in modern voice assistants \citep{chen2019bert}, dialog systems \citep{chen2013unsupervised}, and even text analysis \citep{zhao2023bike}.

\begin{figure}[ht]
\centerline{\includegraphics[width=10cm]{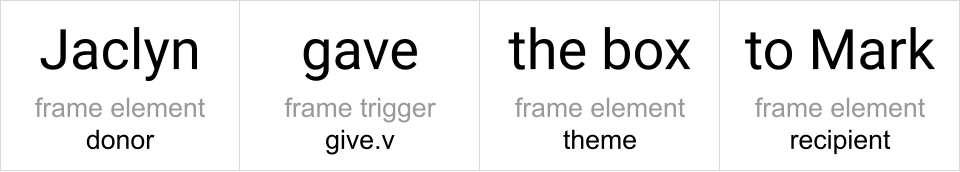}}
    \caption{The sentence "Jaclyn gave the box to Mark" annotated with frame trigger and frame elements for the "Giving" frame.}
    \label{fig:basicframe}
\end{figure}

A semantic frame in FrameNet describes an event, relation, or situation and its participants. When a frame occurs in a sentence, there is typically a  "trigger" word in the sentence which is said to evoke the frame. In addition, a frame contains a list of arguments known as frame elements which describe the semantic roles that pertain to the frame. A sample sentence parsed for frame and frame elements is shown in Figure \ref{fig:basicframe}.

FrameNet provides a list of lexical units (LUs) for each frame, which are word senses with may evoke the frame when they occur in a sentence. For instance, the frame "Attack" has lexical units "ambush.n", "ambush.v", "assault.v", "attack.v", "attack.n", "bomb.v", and many others. These lexical units are not exhaustive, however - a frame trigger may not necessarily be one of the lexical units listed in the frame, but the lexical units provide a strong hint that the frame may be present.

In this paper we treat frame semantic parsing as sequence-to-sequence text generation task, and fine-tune a T5 transformer model \citep{raffel2020exploring} as the base model. We increase performance by pretraining on related datasets, providing in-context prompt hints to T5 based on FrameNet data, and using textual data augmentations \citep{ma2019nlpaug} to increase training data. More details on our implementation are given in Section \ref{sec:method}

We evaluate the performance of Frame Semantic Transformer using the same dataset splits as Open Sesame \citep{swayamdipta2017frame}, the previous state-of-the-art open-source parser. Frame Semantic Transformer exceeds the performance of Open Sesame, and achieves near state-of-the-art performance compared with modern frame semantic parsers that do not publish models \citep{kalyanpur2020open,zheng2022double}.

The performance of Frame Semantic Transformer does not, however, come at the cost of usability. The library can be installed via PyPI \citep{pypi} with the following command:

\begin{verbatim}
pip install frame-semantic-transformer
\end{verbatim}

Performing frame semantic parsing on a sentence can be achieved with a few lines of code. We leverage the Huggingface \citep{wolf-etal-2020-transformers} model hub and NLTK \citep{bird2009natural} corpora so that all required models and datasets are automatically downloaded when frame semantic transformer is first run, requiring no further action by the user aside from installing the library from PyPI. Basic usage is shown in Figure \ref{fig:basic_usage}. Pretrained models are provided based on T5-base and T5-small, with T5-base being the default model used. The code for Frame Semantic Transformer is available on Github \footnote{\href{https://github.com/chanind/frame-semantic-transformer}{https://github.com/chanind/frame-semantic-transformer}}.

\begin{figure}[ht]
    \centering
    \begin{verbatim}
        from frame_semantic_transformer import FrameSemanticTransformer
        frame_transformer = FrameSemanticTransformer()
        sentence = "The hallway smelt of boiled cabbage and old rag mats."
        results = frame_transformer.detect_frames(sentence)
    \end{verbatim}
    \caption{Performing frame semantic parsing requires only a few lines of code using Frame Semantic Tranformer. All needed pretrained models and datasets are downloaded automatically.}
    \label{fig:basic_usage}
\end{figure}

\section{Method}
\label{sec:method}

Typically, frame semantic parsing approaches treat the task as a set of 3 subtasks which happen in serial \citep{kalyanpur2020open}. First, in the trigger identification subtask, all trigger locations are identified in the text where a frame occurs. Second, in the frame classification subtask, each identified trigger location is classified with a FrameNet frame. Finally, in the arguments extraction subtask, frame elements and their arguments are identified in the text.

We treat each of the 3 subtasks as sequence-to-sequence tasks performed in series by a fine-tuned T5 model. Each of these tasks follows the format \texttt{"<task name> <task-specific hints> : <text>"}.

\subsection{Trigger identification}

Given a sentence, the trigger identification task identifies locations in the sentence text which could be frame triggers. This task is conceptually the simplest of the three - it has no task-specific hints, and the goal of the task is to insert markers in the text to indicate frame triggers. For this task, we us the asterisk character \texttt{*} to indicate a frame trigger. This is shown in Figure \ref{fig:basic_trigger_id}.

\begin{figure}[ht]
    \centering
    \begin{tabular}{rl}
    input: & \texttt{"TRIGGER: It was no use trying the lift."}\\
    output: & \texttt{"It was no use *trying the *lift."}
    \end{tabular}
    \caption{Trigger identification input and expected output for the text "It was no use trying the lift.". Trigger locations are indicated by \texttt{*} in the output}
    \label{fig:basic_trigger_id}
\end{figure}

\subsection{Frame classification}

For each trigger identified in the trigger identification step, a frame classification task is created to classify the frame that the trigger refers to. To make this task easier for the model, we use the LUs from each frame to build a list of possible frames this trigger could refer to.

We normalize trigger words and frame LUs using a similar process. First, we lowercase the word and stem and lemmatize it using multiple stemmers and lemmatizers. Each stemmer and lemmatizer may treat different English words slightly differently, so multiple are used to increase the chance the normalized trigger word will match a normalized LU from framenet. Specifically, we use an English Snowball stemmer \citep{porter1980algorithm}, a Lancaster stemmer \citep{paice1990another}, a Porter stemmer \citep{porter1980algorithm}, and a lemmatizer based on WordNet \citep{miller1995wordnet}, all from NLTK to generate a set of up to 4 possible normalized versions of the trigger word.

For LUs, we also remove the part of speech (POS) tag. T5 is a powerful transformer model and likely does not need to be provided with POS info, although this is something that could be explored in future work. In addition, for the trigger word, we also generate bigrams for the trigger and the words on either side of the trigger, and normalize the bigrams in the same way. Some LUs contain multiple words, so generating bigrams increases the chance that after this normalization process the matching frame is found and can be added as a hint.

For instance, for the trigger word "trying" from Figure \ref{fig:basic_trigger_id}, this word has the bigrams "use\_trying" and "trying\_to", and the monogram "trying". After normalization, these become the lookup set:

\{
    us\_tri, us\_try, us\_trying, use\_tri, use\_try, use\_trying,
    tri, try, trying,
    tri\_to, try\_to, trying\_to
\}

This lookup set overlaps with the normalized LUs for the following frames:

\{
    Attempt,
    Attempt\_means,
    Operational\_testing,
    Tasting,
    Trial,
    Try\_defendant,
    Trying\_out,
\}

finally, these overlapping frames are provided as part of the prompt for the frame classification task as shown below:

\begin{tabular}{rl}
input: & \texttt{``FRAME Attempt Attempt\_means Operational\_testing Tasting Trial}\\
       & \hspace{5mm} \texttt{Try\_defendant Trying\_out: It was no use *trying the lift.''}\\
output: & \texttt{``Attempt\_means''}
\end{tabular}

Likewise, a frame classification task is generated for the trigger ``lift" as well, as shown below:

\begin{tabular}{rl}
input: & \texttt{``FRAME Body\_movement Building\_subparts Cause\_motion Cause\_to\_end}\\
       & \hspace{5mm} \texttt{Connecting\_architecture Theft: It was no use trying the *lift.''}\\
output: & \texttt{``Connecting\_architecture''}
\end{tabular}

\subsection{Argument extraction}

After a frame is identified, the next task is to identify the frame elements and arguments for that frame in the text. An argument extraction task is generated for every frame classified. We include all available frame element names from FrameNet for the frame in question as part of the prompt input to make the argument extraction task easier for T5. The output is of the form \texttt{``<element 1>="<arguments 1>" | <element 2>="<arguments 2>" | \ldots''}. For instance, the arguments extraction task for the \texttt{Attempt\_means} frame from above is shown below:

\begin{tabular}{rl}
input: & \texttt{``ARGS Attempt\_means | Agent Means Goal Circumstances Degree Depictive Domain }\\
       & \hspace{5mm} \texttt{Duration Frequency Manner Outcome Particular\_iteration Place Purpose}\\
       & \hspace{5mm} \texttt{Time: It was no use *trying the lift.''}\\
output: & \texttt{``Means="the lift"''}
\end{tabular}

Likewise, the arguments extraction task for the \texttt{Connecting\_architecture} frame is shown below:

\begin{tabular}{rl}
input: & \texttt{``ARGS Connecting\_architecture | Part Connected\_locations Creator }\\
       & \hspace{5mm} \texttt{Descriptor Direction Goal Material Orientation Source Whole: }\\
       & \hspace{5mm} \texttt{It was no use trying the *lift.''}\\
output: & \texttt{``Part="the lift"''}
\end{tabular}

\subsection{Pretraining}

The training data for FrameNet 1.7 is relatively small with under 6,000 fully annotated sentences total, so it is common to leverage FrameNet exemplar data as well to increase the amount of training data available. This exemplar data includes around 100,000 sentences. Exemplar sentence annotates only a single frame per sentence, so it is not suitable for generating trigger identification tasks, but it is still a rich source of data to improve performance on frame classification and argument extraction tasks. The distribution of exemplar sentences is different from the distribution of training data for FrameNet 1.7 \citep{kshirsagar2015frame}, so rather than train on exemplar data directly we instead use it for pretraining.

Another rich source of additional training data is PropBank \citep{kingsbury2002treebank}. PropBank is a similar frame parsing dataset to FrameNet, although PropBank tends to focus more on verbs than FrameNet and has simpler arguments. Still, the tasks are similar enough that pretraining on PropBank can help the model score higher on FrameNet 1.7. Specifically, we use the PropBank training data from OntoNotes \citep{weischedel2013ontonotes} and the English Web Treebank \citep{ewt}.

During training, we begin with a pretrained T5 model from Huggingface \citep{wolf-etal-2020-transformers}, and then go through two additional iterations of pretraining. First we pretrain on PropBank data, and then on FrameNet 1.7 exemplars, before finally training on the FrameNet 1.7 training set.

\subsection{Data augmentation}

The FrameNet 1.7 training data is well formatted and grammatically correct, but in reality a lot of text that needs to be semantically parsed is not well formatted and may have errors and typos. To help make our model more robust to real world data, we also use data augmentation to expose the model to misspellings, synonyms, and other differently formatted sentences.

The textual augmentations used are the following, leveraging the nlpaug Python library \citep{ma2019nlpaug}:
\begin{itemize}
  \item Synonyms: swaps out a word for a synonym from WordNet \citep{miller1995wordnet}.
  \item Quotations: replaces latex-style quotes with standard double quotes and vice-versa.
  \item Random misspelling: replaces characters in words with different characters at random.
  \item Keyboard misspelling: replace characters with typos likely based on key locations on keyboards.
  \item Uppercase and lowercase: fully uppercase or lowercase the sentence.
  \item Delete punctuation: randomly deletes punctuation characters in the sentence.
\end{itemize}

When augmenting text during training, we make sure to adjust the indices of triggers and frame elements to match the new locations after the augmentation is applied.

\subsection{Task balancing}

There is a mismatch between the 3 subtasks in terms of how much training data each task has. Frame classification and argument extraction have multiple examples per training sentence, since a task is generated for every frame in a sentence. Furthermore, these tasks also benefit from pretraining with FrameNet exemplar data. Trigger identification, however, has only 1 training example per sentence, and cannot learn from exemplar data, so there is a large mismatch between the amount of trigger identification samples available and the amount of frame classification and argument extraction samples.

To help address this, we sample trigger identification tasks at a 3x higher rate than frame classification and argument extraction tasks during training to help ensure that the trigger identification performance does not trail behind that of the other tasks per training epoch. We also increase the data augmentation rate for trigger identification tasks to help increase the number of training samples available.

\section{Evaluation}

FrameNet 1.7 does not include an official train / test / dev split, so we follow the split and evaluation used by Open Sesame \citep{swayamdipta2017frame} as this is the most popular open-source frame semantic parser on Github, and was also the previous state-of-the-art. Other parsers also use the same split for this reason \citep{kalyanpur2020open,zheng2022double}.

We calculate f1 score for each of the subtasks against the dev and test sets from Open Sesame. For trigger identification, each trigger location that is identified correctly is considered a true positive, each location that is missed is a false negative, and each location that is incorrectly marked is a false positive. For frame classification, an incorrectly classified frame is considered both a false positive and a false negative. For the argument extraction task, each frame element that is correctly identified and labeled is a true positive. If a frame element is missed it is a false negative. If a frame element is marked incorrectly it is a false positive. If a frame element is labeled, but the element is incorrectly classified or the arguments are not labeled entirely correctly, this is considered both a false positive and a false negative.

\begin{figure}[ht]
    \centering 
    \begin{tabular}{ |p{6cm}|p{0.9cm}|p{0.9cm}|p{0.9cm}|p{0.9cm}|p{0.9cm}|p{0.9cm}|  }
     \hline
     \thead{Model} & \multicolumn{3}{c|}{\thead{FN 1.7 dev set}} & \multicolumn{3}{c|}{\thead{FN 1.7 test set}}\\
     & \multicolumn{1}{c}{Trigger ID} & \multicolumn{1}{c}{Frame ID} & \multicolumn{1}{c|}{Args ID}
     & \multicolumn{1}{c}{Trigger ID} & \multicolumn{1}{c}{Frame ID} & \multicolumn{1}{c|}{Args ID}\\
     \hline
     \citep{peng2018learning} & - & - & - & - & \textbf{0.891} & - \\
     \citep{zheng2022double} & - & - & - & - & - & 0.756\\
     \citep{kalyanpur2020open} & - & - & \textbf{0.77} & - & - & \textbf{0.76} \\
     \hline
     Open Sesame \citep{swayamdipta2017frame} & \textbf{0.80} & 0.90 & 0.61 & 0.73 & 0.87 & 0.61\\
     Frame Semantic Transformer (T5-small)    & 0.75  & 0.87  & 0.76 & 0.71 & 0.86 & 0.73\\
     Frame Semantic Transformer (T5-base)    & 0.78  & \textbf{0.91}  & \textbf{0.78} & \textbf{0.74} & \textbf{0.89} & \textbf{0.75}\\
     \hline
    \end{tabular}
    \caption{Evaluation results comparing Frame Semantic Transformer with other frame semantic parsers on FrameNet 1.7, where comparable data can be found. The top section of the table contains frame semantic parsers without available pretrained models, while the bottom section contains open-source parsers with pretrained models. Bold indicates the best performance in each group.}
    \label{fig:results}
\end{figure}

We also include an ablation study showing the effects of pretraining and data augmentation on model performance in Figure \ref{fig:ablation}. Data augmentation actually slighltly hurts performance in argument extraction, but we still think it is worth it to give the model more robustness to messy examples that may appear in real-world data. Lack of pretraining appears to slightly harm performance on all tasks. We did not do a statistical significance test on this data.

\begin{figure}[ht]
    \centering 
    \begin{tabular}{ |p{6cm}|p{1.5cm}|p{1.5cm}|p{1.5cm}| }
     \hline
     Model & Trigger ID & Frame ID & Args ID \\
     \hline
     Frame semantic transformer (T5-base) & 0.74 & 0.89 & 0.75 \\
     No data augmentation & 0.74 & 0.89 & 0.76 \\
     No pretrain & 0.72 & 0.88 & 0.74 \\
     \hline
    \end{tabular}
    \caption{Ablation study comparing performance without data augmentation and without pretraining on the FrameNet 1.7 test set.}
    \label{fig:ablation}
\end{figure}

\section{Related work}

Recent work on frame semantic parsing has focused on incorporating more information from FrameNet into the parsing process. \citep{zheng2022double} encode the frame relation graph into an embedding space during inference to improve performance. \citep{su2021knowledge} encodes the full text of frame and element descriptions to aid in classification. Our work also follows in this vein by using lexical unit data to provide hints to our model during frame classification. However, both \citep{zheng2022double} and \citep{su2021knowledge} do not provide pretrained, open-source models.

Most similar to our work, and largely a point of inspiration, is \citep{kalyanpur2020open}. In this work, a T5 model \citep{raffel2020exploring} is fine-tuned on the frame classification and argument extraction tasks. In a variant of their work, the T5 decoder is replaced with a classification head for frame classification. However, this work does not deal with trigger identification, and does not use lexical unit hints during frame classification. Furthermore, no code or models are open-sourced as part of this work, making it difficult for end-users to easily make use of the model.

Previous open-source frame parsers include Open Sesame \citep{swayamdipta2017frame} and SEMAFOR \citep{das2010probabilistic}. However, both of these projects predate the rise of the transformer architecture, and their performance lags behind transformer-based solutions, especially in argument extraction.

\section{Conclusion}

Frame Semantic Transformer approaches or matches state-of-the-art performance on frame semantic parsing tasks while also being easy to use as an end-user. We improve performance by pretraining both on FrameNet exemplars and PropBank data. We also incorporate frame knowledge from FrameNet via lexical units and available frame elements and pass that knowledge to T5 in-context as part of task prompts. In addition, we add NLP data augmentations to help the model generalize to real-world data which will likely be formatted differently than the FrameNet 1.7 training set. At present, Frame Semantic Transformer only provides pretrained models for English FrameNet, but we hope to support other languages and PropBank in the future as well.

\bibliography{references}

\end{document}